\documentclass[11pt,letterpaper]{article}
\usepackage{ijcnlp2017}
\usepackage{times}
\usepackage{latexsym}

\usepackage{times}
\usepackage{helvet}
\usepackage{courier}
\usepackage{amsmath, amssymb}
\usepackage{bm}
\usepackage{float}
\usepackage{graphicx}
\usepackage{caption}
\usepackage{subfig}
\usepackage{multirow}
\usepackage{algorithm}
\usepackage[noend]{algpseudocode}
\usepackage{enumitem}
\usepackage{multirow}

\algnewcommand{\Inputs}[1]{%
  \State \textbf{Inputs:}
  \Statex \hspace*{\algorithmicindent}\parbox[t]{.8\linewidth}{\raggedright #1}
}
\algnewcommand{\Initialize}[1]{%
  \State \textbf{Initialize:}
  \Statex \hspace*{\algorithmicindent}\parbox[t]{.8\linewidth}{\raggedright #1}
}
\newcommand*{\Break}{\textbf{break}}

\ijcnlpfinalcopy

\def\ijcnlppaperid{1138}

\title{Single-Queue Decoding for Neural Machine Translation}

\author{
Raphael Shu, Hideki Nakayama \\
\texttt{shu@nlab.ci.i.u-tokyo.ac.jp, nakayama@ci.i.u-tokyo.ac.jp} \\
\texttt{The University of Tokyo}
}

\date{}

\begin{document}

\maketitle

\begin{abstract}
Neural machine translation models rely on the beam search algorithm for decoding. In practice, we found that the quality of hypotheses in the search space is negatively affected owing to the fixed beam size. To mitigate this problem, we store all hypotheses in a single priority queue and use a universal score function for hypothesis selection. The proposed algorithm is more flexible as the discarded hypotheses can be revisited in a later step. We further design a penalty function to punish the hypotheses that tend to produce a final translation that is much longer or shorter than expected. Despite its simplicity, we show that the proposed decoding algorithm is able to select hypotheses with better qualities and improve the translation performance.
  \end{abstract}

\section{Introduction}

Machine translation models composed of end-to-end neural networks \cite{sutskever2014sequence, bahdanau2014neural,Shazeer2017OutrageouslyLN,Gehring2017ConvolutionalST}  are starting to become mainstream. Essentially, neural machine translation (NMT) models define a probabilistic distribution $p(y_t|y_1, ..., y_{t-1}, X)$ to generate translations, where $X$ is a source sentence, and $y_1, ..., y_{t-1}$ is a history of emitted words for predicting the next word $y_t$. One can think of a NMT model as a neural language model with the source sentence included in the context.

As the search space of possible outputs is incredibly large, we can only afford to explore a limited number of candidates. In practice, we use the beam search algorithm to generate output sequences \cite{Graves2012SequenceTW, sutskever2014sequence}. The algorithm limits the search space by considering only a fixed number of hypotheses (i.e.,~partial translations) in each step, and predicting next words only for the selected hypotheses.

However, we found that the strict limit of hypothesis selection affects the quality of the search space negatively. Since the number of active hypotheses is fixed, the algorithm must give up some existing hypotheses to explore new possible decoding paths, even though the discarded hypotheses are not ``hopeless''. The problem has a similar flavor as the exploration-exploitation dilemma. 

In this work, we extend beam search to introduce more flexibility in hypothesis selection. We manage all discarded hypotheses in a single priority queue so that they can be selected later when necessary. The extended algorithm is guided by a universal score function, which is capable of evaluating the hypotheses of different lengths. To encourage the algorithm to select hypotheses that can potentially result in good final translations, we design a length matching penalty that penalizes the hypotheses that may produce incorrect number of words in the final translation. Experiments show that the proposed algorithm is able to improve the quality of search space and thus results in better translation performance.

%
%

\section{Related Work}



To improve the performance of beam search, a basic technique is length-normalization that simply divides the log-probability by the number of words. As far as we know, it is firstly clearly described in \citet{Graves2012SequenceTW} in the context of recurrent neural networks. We also apply length-normalization in the proposed algorithm.


To improve the quality of the score function in beam search, \citet{Wiseman2016SequencetoSequenceLA} propose to run beam search in the forward pass of training, then apply a new objective function to ensure the gold output does not fall outside the beam. An alternative approach is to correct the scores with reinforcement learning \cite{li2017learning}. This work focuses on fixing the limited search space of beam search rather than the score function.

    \citet{hu2015improved} also describes a priority queue but has a different mechanism and purpose. The priority queue in their work contains top-1 hypotheses from different hypothesis stacks. In each step, only one hypothesis from the queue is allowed to be considered. Their purpose is to use the priority queue to speed up beam search at the cost of performance degradation, which is different to this work. 

\section{Deficiency of Beam Search} \label{sec:problems}

\begin{figure}[t]
  \centering
  \includegraphics[width=0.5\textwidth]{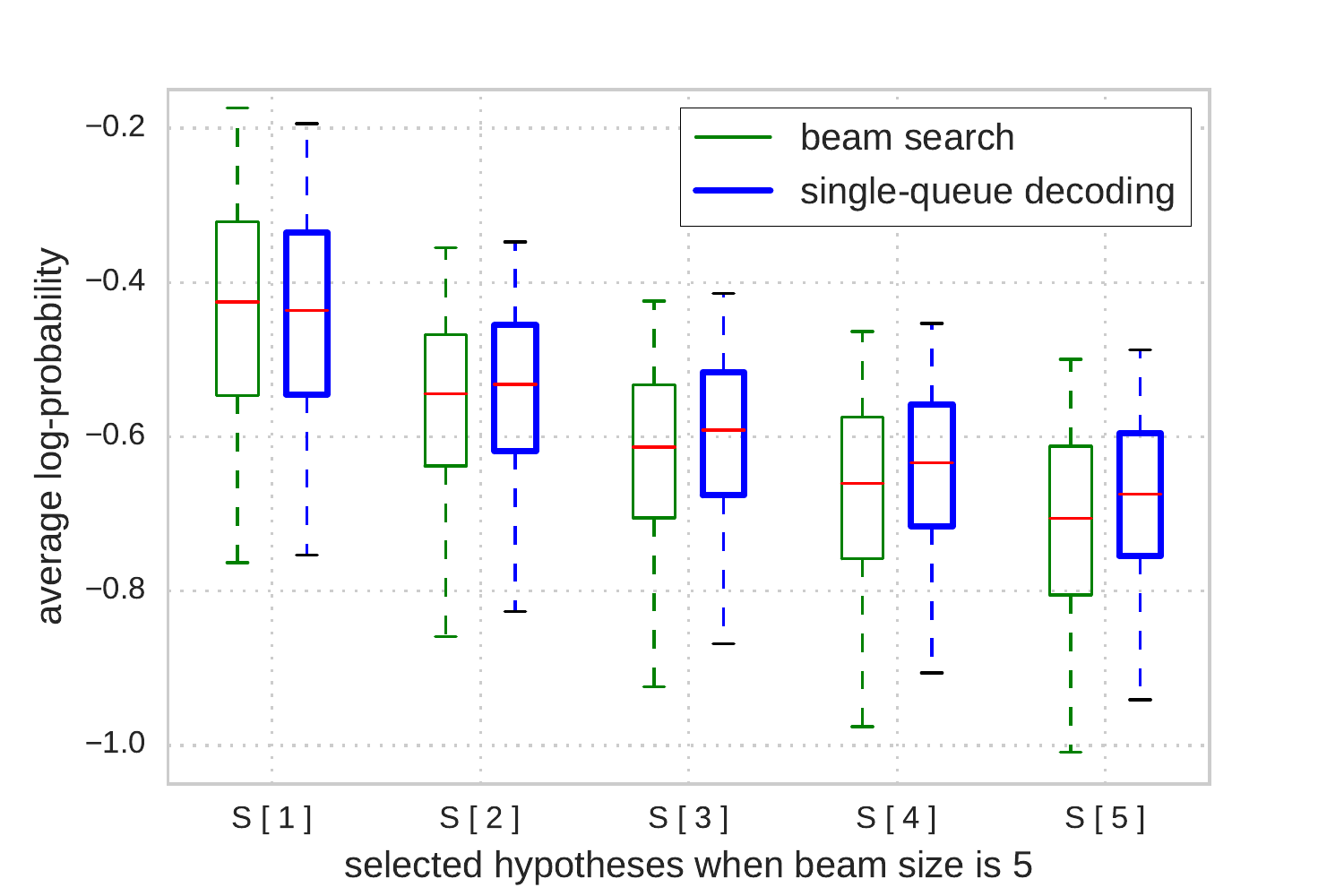}
  \caption{Average scores (log-probability) of the hypotheses in each position of the selected hypothesis list $S$ in each step. $S$ is sorted in descending order. The data is collected by decoding 1k sentences with a beam size of 5. The proposed single-queue decoding is shown to select better alternative hypotheses compared to beam search. }
  \label{fig:fig_rank}
\end{figure}

Beam search finds a hypothesis $y$ that maximizes the log probability $\log p(y|X)$, given an input sentence $X$. In each step, a fixed number of hypotheses are considered by the algorithm. Then the NMT model predicts the probabilities of the next output token for each hypothesis. Suppose that the fixed number (beam size) is $B$, and the vocabulary size is $V$. Then, theoretically, we can obtain a maximum of $B \times V$ new hypotheses. Beam search then keep $B$ hypotheses with highest log probabilities. Thus, the hypotheses considered in each step have exactly the same length (i.e.,~number of tokens). The algorithm ends when $B$ finished translations are collected.


Since the beam size is fixed, when the algorithm attempts to explore multiple new decoding paths for a hypothesis, it has to discard some existing decoding paths. However, the new decoding paths may be found to have low scores in near future. Because the discarded hypotheses can not be revisited, beam search has to continue searching in a low-quality decoding paths. As a result, some hypotheses in the beam may have much lower qualities. Fig.~\ref{fig:fig_rank} shows the average log-probabilities (normalized by length) of the hypotheses in each position of the beam, which indicates that the quality of alternative hypotheses is significantly worse than the best hypothesis. One can also think this problem as a result of limited search space, where the path from the ``BOS'' token to the ``EOS'' token is limited in width.


\section{Single-Queue Decoding}

In this section, we introduce an extended version of beam search, which maintains a single priority queue that contains all possible hypotheses. We refer to this algorithm as {\it single-queue decoding} in this paper. In contrast to the standard beam search, which only considers hypotheses with same length in each step, the proposed algorithm can select arbitrary hypotheses that differ in length.

By allowing mixing hypotheses of different lengths, the proposed algorithm is able to ``regret'' its decisions and explore a discarded hypotheses if the front (i.e.,~longer) hypotheses have worse scores.

\subsection{Main Algorithm}

The pseudo code of single-queue decoding algorithm is shown in Alg.~\ref{alg:decoding}. Let $B$ be the beam size, and $T$ be the number of max steps. Similar to the standard beam search, we run the decoding algorithm until $B$ finished translations are collected. In the worst case, the algorithm will run for a maximum of $T$ steps. All collected hypotheses are maintained in a priority queue $H$.

\begin{algorithm}[t]
\caption{Single-queue decoding}\label{alg:decoding}
\begin{algorithmic}
\Initialize{
    $B \gets$ beam size \\
    $H \gets$ empty hypothesis queue \\
    $T \gets $ max steps \\
}
\vskip 2pt
\For{$t \gets 1 \textrm{ to } T$}
    \State $S \gets $ select best $B$ unfinished hyps in $H$
    \State Remove hyps in $S$ from $H$
    \State $S^\prime \gets$ decode $S$ to get $B \times B$ new hyps with highest local scores
    \State Evaluate hyps in $S^\prime$ with Eq. \ref{eq:score}
    \State Mark hyps end with a EOS token as finished
    \State $\hat S \gets$ select best $2\small{\times}B$ hyps in $S^\prime$
    \State Merge $\hat S$ into $H$
    \If{$\#( \mbox{finished hyps in } H) \geq B$}
    \State \Break
    \EndIf
\EndFor
\State $\hat y \gets$ best finished hyp in $H$
\State \textbf{output} $\hat y$
\end{algorithmic}
\end{algorithm}

The proposed decoding algorithm relies on an universal score function $\mathrm{score}(y)$ to evaluate a hypothesis $y$. In each step, hypotheses with the highest scores are removed from the queue to predict next words for them. We collect $B \times B$ hypotheses that have top local scores. Specifically, for a hypothesis $y_{1:l}$\footnote{Note that $y_{1:l}$ is a short form of $y_1, ..., y_l$, which is a partial translation with $l$ words.}, we keep $B$ predictions $y^{(1)}_{l+1}, ..., y^{(B)}_{l+1}$ with highest probabilities. This simple filtering can avoid huge computational cost caused by the score function. We mark all hypotheses end with a ``EOS'' token as finished, so that they will not be decoded anymore. Then, we evaluate these new hypotheses with a universal score function, and retain the top $2 \times B$ hypotheses in a list $\hat S$. Finally, $\hat S$ is merged into the queue $H$. Note that, if we keep only $B$ hypotheses in $\hat S$, then the algorithm will produce the same result as beam search.

\subsection{Universal Score Function}

The universal score function for evaluating a hypothesis in the queue has the following form:
\begin{equation}
    \label{eq:score}
    \mathrm{score}(y) = \frac{1}{{|y|}^\lambda} \log p(y|X)  + \mathrm{PG}(y) + \mathrm{LMP}(y).
\end{equation}

The first part of the equation is the log probability with length-normalization, where $\lambda$ is a hyperparameter that is similar to the definition of length penalty in \citet{wu2016google}. The second part of Eq. \ref{eq:score} is a progress penalty, which encourages the algorithm to select longer hypotheses:
\begin{equation}
    \mathrm{PG}(y) =
    \begin{cases}
        0 & \text{if } y_{1:l} \text{ finished} \\
        \alpha \frac{|y|^\beta}{|X|^\beta} & \text{otherwise}
    \end{cases}
\end{equation}
where $\alpha$ and $\beta$ are the weights that control the strength of this function. The progress penalty is crucial to single-queue decoding, as it ensures that the decoding algorithm is progressing in general.

The last part of Eq.~\ref{eq:score} is a length matching penalty, which will be described in Section \ref{sec:lmp}.

\subsection{Length Matching Penalty}
\label{sec:lmp}

The standard beam search evaluates hypotheses solely based on the emitting probabilities, which may result in a final translation much longer or shorter than expected. To compensate for this deficiency, we design a length matching penalty. The intuition is to correct the scores by predicting a Gaussian distribution of correct translation length, and penalize all hypotheses that tend to produce much longer or shorter translations.

Let the first state of the backward encoder in a standard NMT model \cite{bahdanau2014neural} be $\bar h_0$. We predict the mean and variance of the distribution of correct translation length using a simple neural network $f^e$:
\begin{align}
    v^e &= f^e(\bar h_0; \theta_e), \\
    \mu^e &= v^e[0];\ \sigma^e = \mathrm{softplus}(v^e[1]),
\end{align}
where, $v^e$ is a two-dimensional vector. To predict the distribution of the final length for a hypothesis $y_{1:l}$, we use a tiny LSTM \cite{hochreiter1997long} followed by a simple neural net $f^d$:
\begin{align}
    h_l &= \mathrm{LSTM^d}(e(y_{1:l});\theta_d) \\
    v^d_l &= f^d(h_l + \bar h_0;\theta_d), \\
    \mu^d_l &= v^d_l[0];\ \sigma^d_l = \mathrm{softplus}(v^d_l[1]),
\end{align}
where, $e(\cdot)$ represents the embeddings of tokens. We train the parameters $\theta_e$ and $\theta_d$ with fixed NMT parameters. Let $L^*$ be the length of the gold output, $L$ be the length of a sampled output obtained by greedy decoding, the loss function is defined as
\begin{align}
    J = - \log P(L^*;\mu^e,\sigma^e) - \frac{1}{L} \sum_{l=1}^{L} \log P(L;\mu^d_l,\sigma^d_l)
\end{align}
where $P(\cdot)$ is a Gaussian distribution with the specified mean and variance.

Finally, the length matching penalty of a hypothesis $y_{1:l}$ is given by
\begin{align}
    \mathrm{LMP}(y_{1:l}) =
    \begin{cases}
        0
        \:\:\:\:\:\:\:\:\:\:\:\:\:\:\:\:\:\:\:\:\:\:\:
        \:\:\:\:\:\:\:\:\:\:
        \text{if } y_{1:l}  \text{ finished} \\
        \gamma * \mathrm{I}(\mathrm{LMS}(y_{1:l}) > \tau)
        \:\:\:\:
        \text{otherwise}
    \end{cases}
    \label{eq:lmp}
\end{align}
where $\mathrm{I}(\cdot)$ is an indicator, $\gamma$ and $\tau$ are hyperparameters. $\mathrm{LMS(\cdot)}$ computes a length matching score:
\begin{align}
    & \mathrm{LMS}(y_{1:l}) = {\mathbb{E}}_{x \sim P(x;\mu^d_l,\sigma^d_l)} \Bigl [ - \log P(x;\mu^e,\sigma^e) \Bigr ]
    \label{eq:lms}
\end{align}

The LMS function outputs a large value when the final length tends to differ from the correct length. Note that Eq.~\ref{eq:lms} is a cross-entropy of two Gaussians, which can be deterministically computed as $\frac{1}{2} \log (2 \pi {\sigma^d_l}^2) + \frac{{\sigma^e}^2 + ({\mu^d_l} - {\mu^e})^2}{2 {\sigma^d_l}^2}$.


\section{Experiments}

\subsection{Experimental Settings}

We evaluate the proposed decoding algorithm with an off-the-shelf NMT model \cite{bahdanau2014neural}. The embeddings and LSTM layers have 1000 hidden units. We evaluate the algorithms on ASPEC English-Japanese translation dataset \cite{NAKAZAWA16.621}. We keep a 80k vocabulary for English side and 40k vocabulary for Japanese side. The NMT model is trained with Adam \cite{kingma2014adam} for 6 epochs using a learning rate of $0.0001$. We report BLEU score based on a standard post-processing procedure \footnote{We use Kytea to re-tokenize results. Details can be found in \url{http://lotus.kuee.kyoto-u.ac.jp/WAT/}.}. 

The additional network components for computing length matching penalty ($f^e$, $f^d$ and $\mathrm{LSTM}^d$) have 128 hidden units in our experiments. They are trained with Adam for 2 epochs with the NMT parameters fixed.

The hyperparameters of the decoding algorithms are tuned by Bayesian optimization \cite{Snoek2012PracticalBO} on a small validation set composed of 500 sentences. We focus on evaluating algorithms with a small beam size, which is more useful in a productive system. We allow the decoding algorithms to run for a maximum of 150 steps.

\subsection{Evaluation Results}

The main results are shown in Table \ref{table:enja}, which use a beam size of 5. The results show that the proposed single-queue decoding (SQD) algorithm significantly improves the quality of translations. With the length matching penalty (LMP), SQD outperforms beam search with length-normalization by 1.14 BLEU on test set. Without the progress penalty (PG), the scores are much worse.

Since SQD computes $B$ hypotheses in batch mode in each step just like beam search, the computational cost inside the loop of Alg.~\ref{alg:decoding} remains the same. The factor affecting the computational cost is the actual number of decoding steps. To clarify that SQD does not improve the performance by significantly increasing the number of steps, we also report the average number of steps and decoding time for translating one sentence in the right-most columns.
We can see that the average number of steps that SQD computes is still close to beam search.
Note that our implementation for LMP is not fully optimized, there is room for further speed improvement.

\newcommand{\specialcell}[2][c]{%
  \begin{tabular}[#1]{@{}c@{}}#2\end{tabular}}

\begin{table}[t]
  \begin{center}
    \begin{tabular}{r|c|c|c|c} \hline \hline
      \small{} & \multicolumn{2}{c|}{BLEU(\%)} & \multirow{2}{*}{\small{\#step}} & \multirow{2}{*}{\small{\specialcell{time\\(ms)}}} \\
      \cline{2-3}
      \small{} &  \small{valid} &  \small{test} &   & \\
      \hline
      \small{vanilla beam search} & 29.61 & 32.87 & 30.3 & 199 \\
      \small{w/ length-norm} & 37.16 & 34.29 & 30.3 & 208  \\
      \hline
      \small{SQD w/o PG, LMP} & 38.09 & 34.62 & 36.1 & 238 \\
      \hline
      \small{SQD w/ PG} & 38.50 & 35.03 & 33.8 & 225 \\
      \small{SQD w/ PG, LMP} & {\bf 38.93} & {\bf 35.43} & 35.0 & 260 \\
      \hline \hline
    \end{tabular}
    \caption{Evaluation results on ASPEC En-Ja task with a beam size of 5}
    \label{table:enja}
  \end{center}
\end{table}

In order to get insight into how SQD improves the performance, we plot the average log-probability of the selected hypotheses in Fig.~\ref{fig:fig_rank}. We can see that SQD improves the quality of the hypotheses other than the first one in the beam, which indicates that the proposed algorithm can rescue high-quality hypotheses from the queue that are previously discarded. 

\section{Conclusion}

In this paper, we extend the beam search with a single hypotheses queue, which can revisit a discard hypothesis, and thus more flexible in decoding. We design a length matching penalty to further help the proposed algorithm to select a hypothesis that can potentially produce a final translation with correct length.

Although the proposed algorithm does not cause a speed issue, it requires a block of GPU memory for storing the decoder states of discarded hypotheses, which has a shape of $T \times B \times M$, where $T$ is the maximum steps, $B$ is the beam size and $M$ is the dimension of the LSTM states. The increased memory usage does not cause a problem unless $T$ and $B$ are both large numbers.

The proposed algorithm is still compatible with other techniques, such as the threshold-based pruning method \cite{freitag2017beam}, reinforcement learning based scoring \cite{li2017learning}, reducing Softmax computation \cite{hu2015improved, l2016vocabulary} and diverse decoding \cite{Li2016ASF,Li2016MutualIA}.

\bibliography{mybib}
\bibliographystyle{ijcnlp2017}

\appendix

\section{Supplemental Materials}
\label{sec:supplemental}

In this work, we focus on testing the performance of our proposed algorithm with a small beam size. Theoretically, one can alleviate the problem of limited search space by using a very large beam size. However, the increased computational cost makes it impractical in a productive system. As supplemental data, we also report the experiment results with different beam sizes in Table \ref{table:beam_size}.

\begin{table}[h]
  \begin{center}
    \begin{tabular}{r|c|c|c} \hline \hline
      \small{} & \multicolumn{3}{c}{Test BLEU(\%)} \\
      \cline{2-4}
      \small{} &  \small{BS=5} &  \small{BS=8} & \small{BS=12}  \\
      \hline
      \small{vanilla beam search} & 32.87 & 32.91 & 32.67 \\
      \small{w/ length-normalization} & 34.29 & 34.82 & 35.05  \\
      \hline
      \small{SQD w/ PG} & 35.03 & 35.44 & 35.65 \\
      \small{SQD w/ PG, LMP} & {\bf 35.43} & {\bf 35.54} & {\bf 35.75} \\
      \hline \hline
    \end{tabular}
    \caption{Evaluation results on ASPEC En-Ja task with different beam sizes (BS).}
    \label{table:beam_size}
  \end{center}
\end{table}

Our implementation is based on Theano. We utilize the Python package ``bayes\_opt'' for Bayesian optimization. We apply the default acquisition function ``ucb'' with a $\kappa$ value of 5. The hyperparameters of the length matching penalty are searched independently from others. We first explore 20 initial points, then optimize for another 20 iterations.
The source code of this work along with a toolkit that allows one to apply single-queue decoding and test new penalty functions for any encoder-decoder model, will be open-sourced in \url{https://github.com/zomux/nmtdec}.

\end{document}